\pdfoutput=1

\documentclass[11pt]{article}

\usepackage[final]{coling}

\usepackage{times}
\usepackage{latexsym}

\usepackage[T1]{fontenc}

\usepackage[utf8]{inputenc}

\usepackage{microtype}

\usepackage{inconsolata}

\usepackage{graphicx}
\newcommand{\ignore}[1]{}

\title{Streamlining Biomedical Research with Specialized LLMs}

\author{Linqing Chen, Weilei Wang, Yubin Xia, Wentao Wu, Peng Xu, ZilongBai, \\Jie Fang, Chaobo Xu, Ran Hu, Licong Xu, Haoran Hua, Jing Sun, Hanmeng Zhong, Jin Liu,  \\ Tian Qiu, Haowen Liu, Meng Hu, Xiuwen Li, Fei Gao, Yong Gu, Tao Shi, Chaochao Wang, \\Jianping Lu, Cheng Sun, Yixin Wang, Shengjie Yang, Yuancheng Li, Lu Jin, Lisha Zhang, \\Fu Bian, Zhongkai Ye, Lidong Pei, Changyang Tu\\
        PatSnap Co., LTD. Suzhou, China\\
\{chenlinqing, wangweilei\}@patsnap.com}

\begin{document}
\maketitle
\begin{abstract}
In this paper, we propose a novel system that integrates state-of-the-art, domain-specific large language models with advanced information retrieval techniques to deliver comprehensive and context-aware responses. Our approach facilitates seamless interaction among diverse components, enabling cross-validation of outputs to produce accurate, high-quality responses enriched with relevant data, images, tables, and other modalities. We demonstrate the system's capability to enhance response precision by leveraging a robust question-answering model, significantly improving the quality of dialogue generation.
The system provides an accessible platform for real-time, high-fidelity interactions, allowing users to benefit from efficient human-computer interaction, precise retrieval, and simultaneous access to a wide range of literature and data. This dramatically improves the research efficiency of professionals in the biomedical and pharmaceutical domains and facilitates faster, more informed decision-making throughout the R\&D process. Furthermore, the system proposed in this paper is available at \href{https://synapse-chat.patsnap.com}{https://synapse-chat.patsnap.com}.
\end{abstract}

\section{Introduction}

\begin{figure}[!ht]
\begin{center}
\includegraphics[width=2.8in]{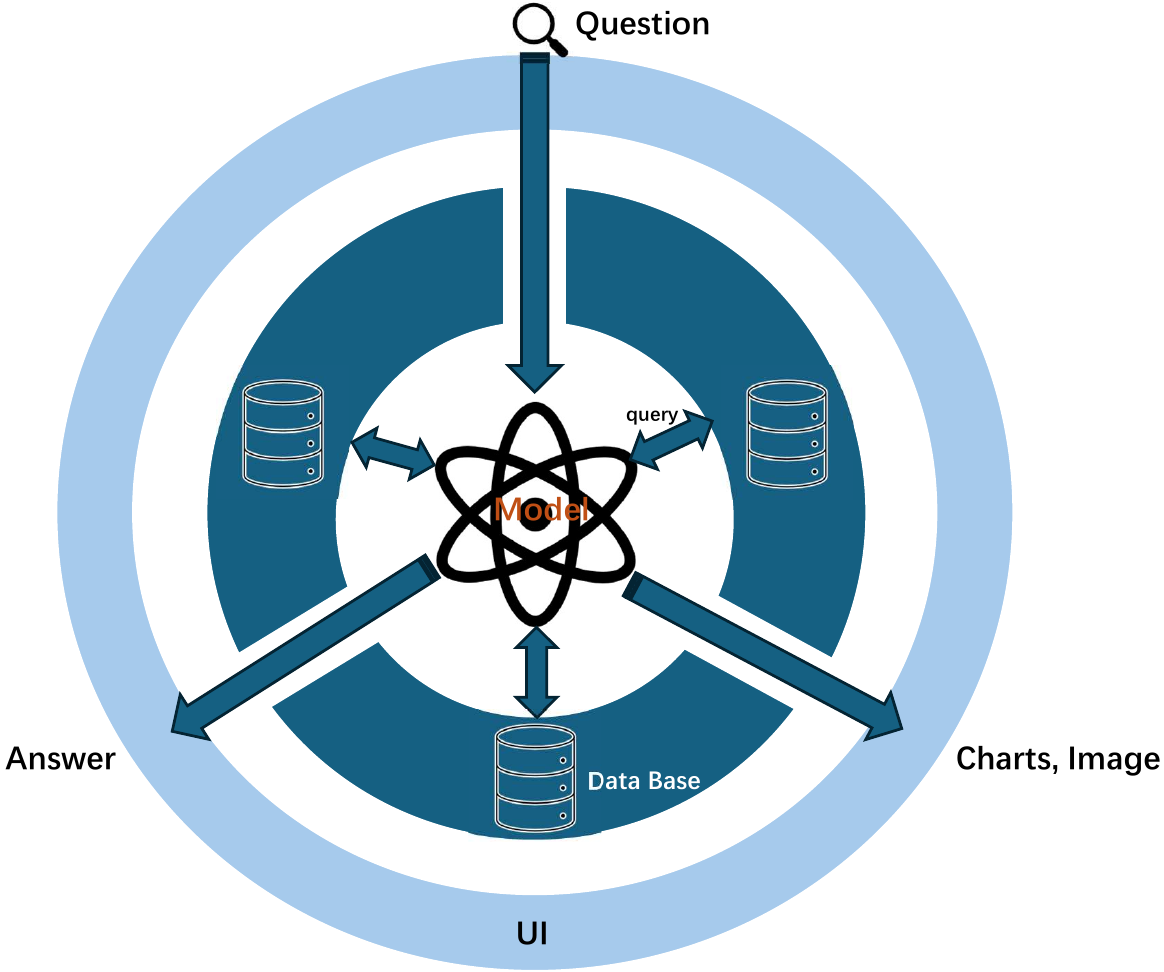}
\end{center}
\caption{Synapse Chat System Architecture} 
\label{fig:over-item}
\end{figure}

The development of Large Language Models (LLMs) has significantly transformed the landscape of natural language processing (NLP). Recent advancements, exemplified by models such as GPT \cite{radford2018improving}, have decreased the reliance on extensive feature engineering, thereby simplifying the creation of complex NLP systems \cite{sarzynska2021detecting, howard2018universal}. These models have demonstrated remarkable capabilities in understanding and generating nuanced text with minimal prompts.
Unlike conventional computational methods, LLMs such as BioBERT \cite{lee2020biobert} and ChemBERTa \cite{chithrananda2020chemberta} excel in deciphering specialized lexicons. Additionally, LLMs have begun integrating data from genomics, proteomics, and chemical databases to provide holistic insights into drug-target interactions \cite{zeng2016convolutional}. For instance, models like Transformer-CNN \cite{karpov2020transformer} illustrate the efficacy of combining LLM architectures with convolutional neural networks to enhance feature extraction in complex datasets.

Recently, we introduced PharmaGPT \cite{chen2024pharmgpt}, the foundational component of the Synapse Chat system. While general-purpose large language models (LLMs) have demonstrated impressive capabilities across a wide range of tasks, their applications in the biopharmaceutical domain have been relatively limited. Existing models often rely on incomplete or narrowly focused datasets, with many emphasizing clinical diagnosis or patient interaction \cite{luo2022biogpt, singhal2023large}. These models lack comprehensive coverage of critical areas such as drug discovery, molecular biology, and regulatory affairs, which are essential for biopharmaceutical research and development. In contrast, PharmaGPT is specifically designed to possess extensive domain knowledge, ensuring full coverage across the biopharmaceutical lifecycle. It integrates advanced capabilities such as natural language to SQL conversion, denoising, and reranking. A key feature of the system is its end-to-end reranking component, which enables the fusion of retrieval results from diverse sources with incomparable scoring mechanisms. This capability facilitates an ensemble approach that incorporates BM25, SQL-based, and vector-based retrieval methods, thereby enhancing the versatility and robustness of the system. Our experimental results demonstrate that PharmaGPT achieves notable improvements in denoising tasks, with relative gains of 2\% to 4\% over prior state-of-the-art general-purpose LLMs.

\begin{figure}[!t]
\begin{center}
\includegraphics[width=3.0in]{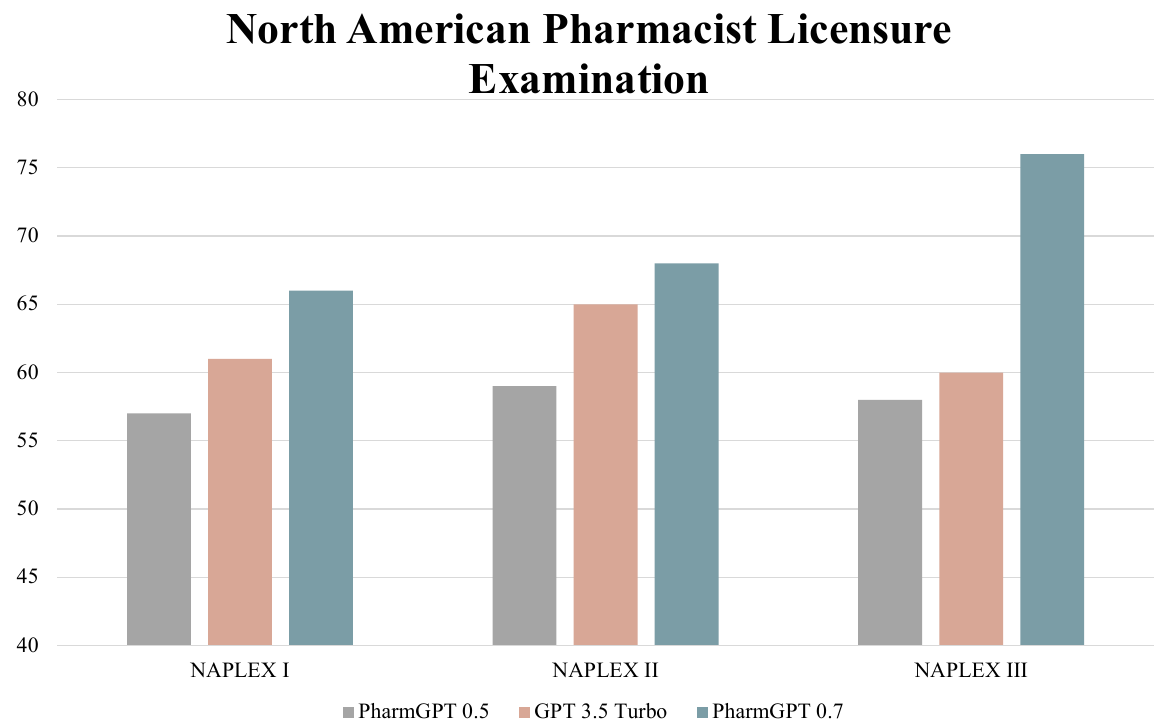}
\end{center}
\caption{Test results of ChatGPT-3.5 Turbo and PharmaGPT models on the North American Pharmacist Licensure Examination (NAPLEX).} 
\label{fig:naplex}
\end{figure}

In this work, we present Synapse Chat, a comprehensive and enhanced system that builds upon the capabilities of PharmaGPT. Synapse Chat supports both asynchronous and real-time user interactions, enabling seamless dialogue, fact-checking, and open-domain question answering. Through extensive empirical evaluations, we demonstrate that Synapse Chat achieves state-of-the-art performance across these tasks within the biopharmaceutical domain. Notably, we introduce a approach that leverages a question-answering model to further enhance dialogue accuracy, significantly improving the system's ability to provide precise and contextually relevant responses.
The Synapse Chat system is designed for a variety of use cases. It allows users to interact with the system at varying levels of verbosity, depending on their specific needs. Additionally, it enables users to cross-examine results across multiple tasks within the same system, providing a holistic and flexible approach to information retrieval and analysis.

\section{System Architecture}

The architecture of the Synapse Chat system, as depicted in Figure \ref{fig:over-item}, is designed to facilitate seamless, real-time, and asynchronous interaction through a web-based interface. At its core, the system integrates multiple components—including a robust user interface (UI), a sophisticated data retrieval engine, and domain-specific large language models (LLMs)—to deliver comprehensive and contextually accurate responses. Users can query the system to retrieve information from various data sources, such as structured databases and unstructured documents, and the system intelligently combines the results through a multi-modal retrieval process. This process leverages an ensemble of retrieval techniques, including traditional keyword-based methods, SQL queries, and advanced vector-based retrieval, ensuring that the most relevant and high-quality information is surfaced. Our PharmaGPT model uses this combined information as input to generate a multi-modal response, assisted by APIs. The final output is then presented to the user in a clear and concise format, which may include text-based answers, charts, or images, depending on the nature of the query.

\subsection{PharmaGPT}

\begin{figure*}[!t]
\begin{center}
\includegraphics[width=6.2in]{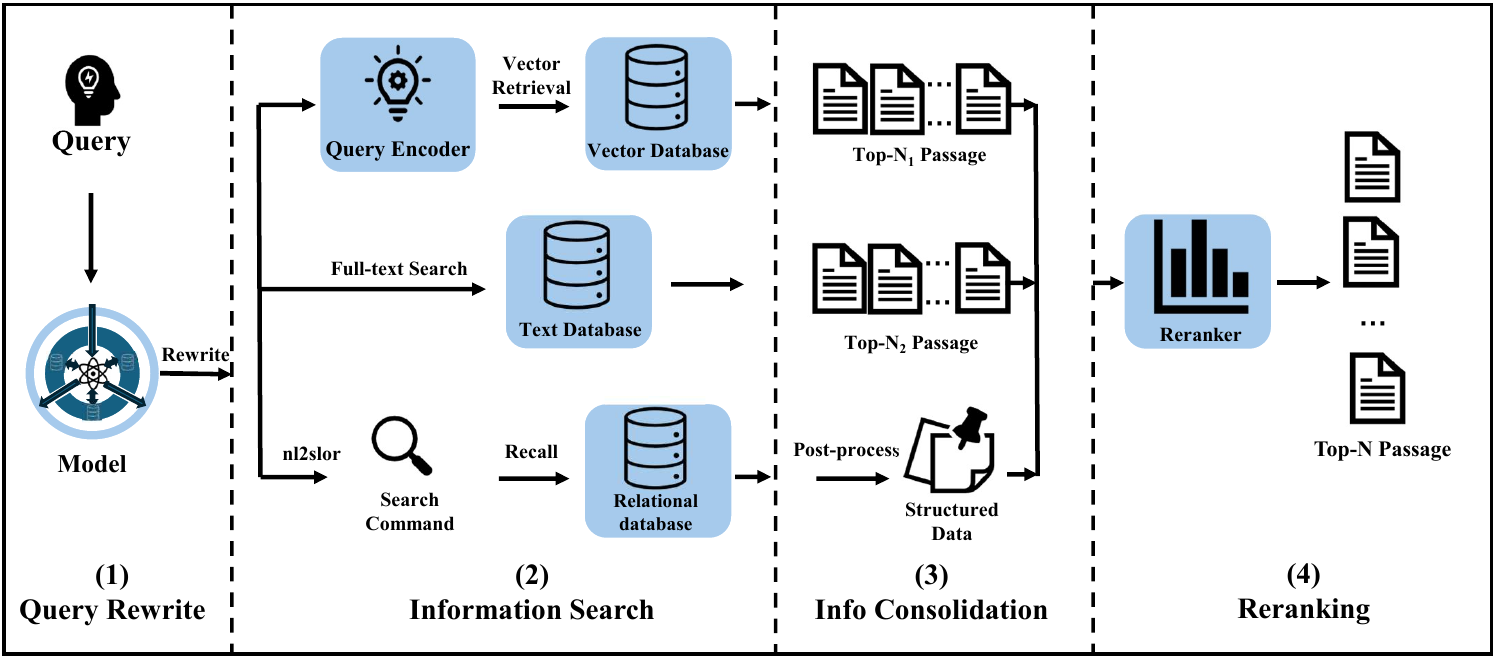}
\end{center}
\caption{Pipeline of the SynapseChat Data Retrieval System} 
\label{fig:retrieval-pipeline}
\end{figure*}

Large language models (LLMs) have significantly transformed Natural Language Processing (NLP) by reducing the reliance on intricate feature engineering. However, their application in highly specialized domains such as biopharmaceuticals and chemistry remains underexplored. These domains are characterized by highly specialized terminologies, complex knowledge structures, and a critical need for precision, areas in which general-purpose LLMs often exhibit limitations. 
In this work, we introduce \textbf{PharmaGPT}, a suite of domain-specific LLMs comprising 13 billion and 70-billion parameter models, meticulously trained on a comprehensive and domain-specific corpus tailored to the biopharmaceutical and chemical sectors. Our evaluation demonstrates that PharmaGPT consistently outperforms existing general-purpose models on domain-specific benchmarks such as the North American Pharmacist Licensure Examination (NAPLEX), showcasing its superior capability in addressing specialized tasks. Notably, this exceptional performance is achieved with models that utilize a fraction—sometimes as little as one-tenth—of the parameters of their general-purpose counterparts.
This breakthrough establishes a new standard for LLMs in the biopharmaceutical and chemical fields, effectively filling the current gap in specialized language modeling. Moreover, it opens new avenues for research and development, paving the way for more accurate and efficient applications of NLP in these highly specialized areas.

Inspired by the work of Angel et al. \cite{angel2023clinical}, we conducted a comprehensive evaluation of our model, PharmaGPT, in comparison with other leading models using the North American Pharmacist Licensure Examination (NAPLEX) dataset. As shown in Fig.~\ref{fig:naplex}, this evaluation not only benchmarks the performance of PharmaGPT in a real-world, domain-specific examination but also highlights its applicability and potential in clinical and pharmaceutical scenarios. PharmaGPT consistently outperforms GPT-3.5 Turbo across several sections of the NAPLEX, underscoring its superior ability to understand and process biopharmaceutical knowledge.

\begin{figure}[!h]
\begin{center}
\includegraphics[width=3.0in]{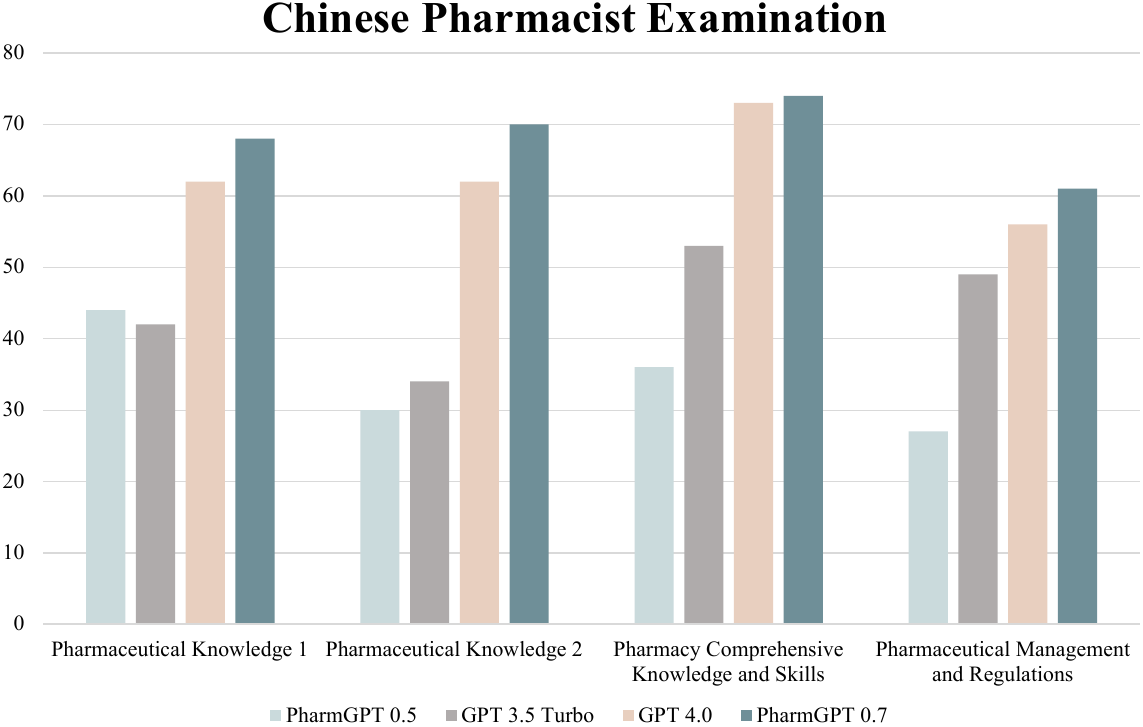}
\end{center}
\caption{Test results of ChatGPT-3.5 Turbo, GPT-4, and PharmaGPT models on the Chinese Pharmacist Examination.} 
\label{fig:chinese}
\end{figure}

As illustrated in Fig.~\ref{fig:chinese}, both versions of PharmaGPT (0.5 and 0.7) demonstrate strong performance across all four categories of the Chinese Pharmacist Examination. Achieving scores in the 70-80\% range, PharmaGPT exhibits robust capabilities in pharmaceutical knowledge, regulations, and comprehensive skills. This consistent high performance indicates that PharmaGPT has been effectively fine-tuned on a large corpus of domain-specific biomedical and pharmaceutical literature, enabling it to excel in regionally and contextually diverse examinations.

\begin{figure}[!h]
\begin{center}
\includegraphics[width=3.3in]{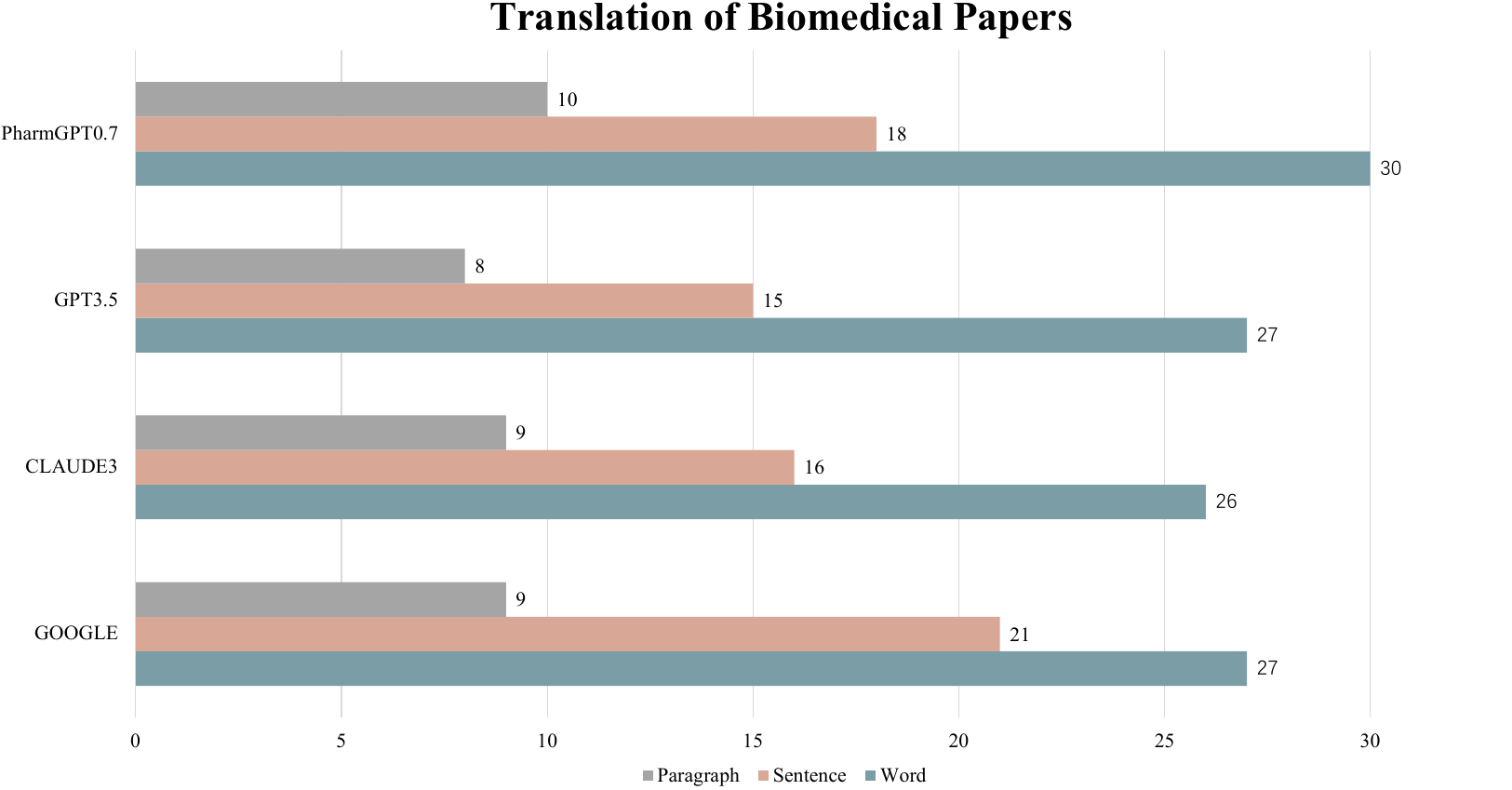}
\end{center}
\caption{Test results of ChatGPT-3.5 Turbo, CLAUDE3, Google Translate, and PharmaGPT models on translation tasks for biomedical papers.} 
\label{fig:translate}
\end{figure}

As shown in Fig.~\ref{fig:translate}, the translation performance of four language models—PharmaGPT 0.7, GPT-3.5, CLAUDE3, and Google Translate—was evaluated across three levels of granularity: paragraph, sentence, and word. Translation quality was measured using BLEU scores~\cite{papineni2002bleu}, with higher scores indicating better performance. PharmaGPT 0.7 demonstrates a clear advantage in translating biomedical papers. At the paragraph level, PharmaGPT 0.7 achieves a BLEU score of 30, outperforming GPT-3.5 (27), CLAUDE3 (26), and Google Translate (27). This trend continues at the word level, where PharmaGPT 0.7 scores 10, compared to GPT-3.5 (8), CLAUDE3 (9), and Google Translate (9). Even at the sentence level, PharmaGPT 0.7 excels with a score of 18, significantly higher than GPT-3.5 (15) and CLAUDE3 (16). These results highlight the superior ability of PharmaGPT 0.7 to handle the complexities of biomedical text translation, making it a highly effective tool for specialized domains.

\subsection{Multi-channel Retrieval System}

To complement the domain-specific large language model (LLM) core tailored for the biopharmaceutical field, we have designed a robust and scalable multi-channel retrieval system. This system is a critical component in achieving our goal of developing a state-of-the-art question-answering platform for specialized biopharmaceutical queries. The retrieval system operates across three distinct data categories: structured data from our proprietary databases, unstructured textual data, and vectorized representations of documents.

As illustrated in Fig.~\ref{fig:retrieval-pipeline}, the unique nature of biopharmaceutical data necessitates a tailored approach to data retrieval. For instance, critical information such as drug development stages, approval statuses, and regulatory filings is stored as structured data in relational databases. This structured data can be efficiently queried using SQL-based methods. Meanwhile, vast corpora of research papers, patents, and clinical trial reports are better suited for vectorization and subsequent retrieval through vector-based search mechanisms. Given the rapid pace of advancements in the biopharmaceutical field, newly acquired or updated textual information is stored as unstructured text segments, which are indexed for full-text search.

The multi-channel retrieval system is designed to seamlessly integrate these three data types. Initially, a query from the user is rewritten by our model in case it is part of a multi-round dialogue. In this situation, our model completes the sentence to facilitate information retrieval. The rewritten query is then processed by the \textbf{Encoder}, which converts it into a suitable format for vector retrieval.
The system performs parallel retrieval using both traditional full-text search for unstructured data and vector retrieval for vectorized documents. For structured data, we use nl2sql (Natural Language to SQL) to convert the user's query from natural language to SQL format. SQL queries are executed to extract relevant results.
The top-N passages from each retrieval channel are combined, and a \textbf{Reranker} is applied to reorder the results based on relevance to the query. This reranked data is subsequently passed to the LLM core as reference material for generating precise and contextually accurate answers.

\subsection{SynapseChat}

As previously discussed, we have developed \textbf{PharmaGPT}, a robust large language model (LLM) specifically tailored for the biopharmaceutical domain. Leveraging its specialized training data, PharmaGPT possesses extensive and in-depth knowledge in biopharmaceuticals, consistently outperforming general-purpose models in domain-specific tasks such as professional examinations and scientific paper translation. PharmaGPT serves as the core "intelligence" of our system, efficiently accessing a vast array of proprietary databases containing biopharmaceutical, chemical, and genetic sequence data. It autonomously ranks, filters, and selects the most relevant information to generate precise, contextually accurate responses while also providing citation references to ensure transparency and reliability. However, to complete the full system, an intuitive and user-friendly interface is essential for facilitating user interaction.

The architecture of the final system is illustrated in Figure~\ref{fig:over-item}. Upon receiving a query from the user, PharmaGPT performs two key functions: (1) it interprets the query, retrieving and processing relevant data from multiple subfields and data types to generate a coherent textual response, and (2) it invokes APIs to dynamically generate visual content such as charts, images, and chemical structure diagrams for compounds, drugs, or gene sequences mentioned in the query. Notably, when the "deep search" mode is activated, PharmaGPT further enhances the output by automatically generating a mind map of the retrieved information, offering users an interactive and comprehensive view of the topic.

\begin{table}[h]
    \centering
    \begin{tabular}{c|c|c|c}
        \hline
        EN & Noise & SynapseChat(ac) & ChatGPT(ac) \\ \hline
         & 0.6 & \textbf{92.33} & 90.00 \\ 
         & 0.8 & \textbf{79.60} & 76.00 \\ \hline
        ZH & Noise &  SynapseChat(ac) & ChatGPT(ac) \\ \hline
         & 0.6 & \textbf{89.33} & 87.67 \\ 
         & 0.8 & \textbf{74.67} & 70.67 \\ \hline
    \end{tabular}
    \caption{Comparative evaluation of SynapseChat and ChatGPT on information relevance discrimination.}
    \label{tab:RGB}
\end{table}

To further assess the system's performance, we conducted a comparative evaluation between \textbf{SynapseChat} and \textbf{ChatGPT} using publicly available datasets \cite{chen2024benchmarking}\footnote{\url{https://github.com/chen700564/RGB/tree/master?tab=readme-ov-file}}. As shown in Table~\ref{tab:RGB}, regardless of the language (Chinese or English) and across different noise levels (0.6 or 0.8), SynapseChat consistently exhibits superior accuracy in discerning relevant information from irrelevant data. This enhanced ability to filter and utilize retrieved information effectively enables SynapseChat to provide users with more accurate and contextually appropriate responses, ultimately resulting in a better user experience. These findings underscore the robustness of SynapseChat in handling noisy data, which is a critical feature for real-world applications in the biopharmaceutical domain.

\section{Applications for Enhancing Research}

\subsection{Domain-Specific Question Answering}

\begin{figure}[!h]
\begin{center}
\includegraphics[width=3.0in]{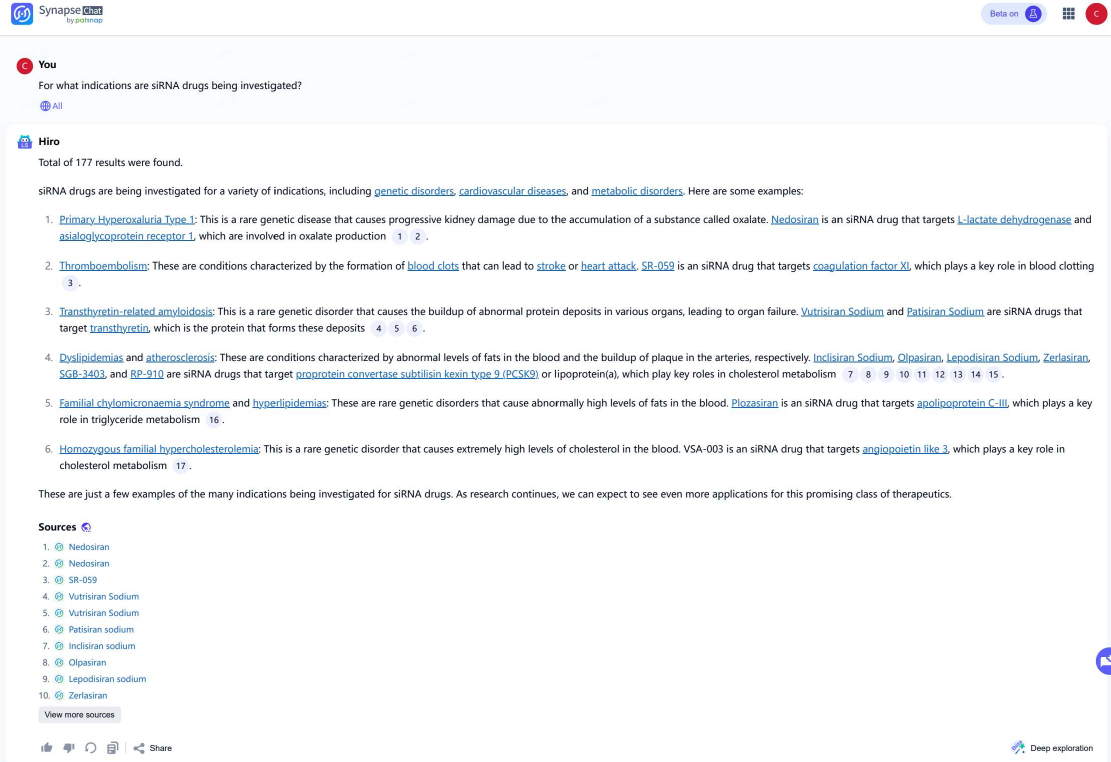}
\end{center}
\caption{Drug-related Question Answering} 
\label{fig:drug}
\end{figure}

In Figures~\ref{fig:drug}, \ref{fig:clinical}, \ref{fig:compound}, and \ref{fig:FDA}, we demonstrate that our system integrates comprehensive and specialized databases encompassing drugs, clinical trial reports, chemical compounds, and regulatory information. This extensive data foundation enables the system to perform highly accurate, domain-specific question answering (QA) across these areas. The system not only retrieves precise answers but also provides detailed references, allowing users to navigate directly to the source material for further exploration. This capability ensures transparency and reliability in the information presented, which is essential for research purposes.

\subsection{Compound Structure Recognition}

\begin{figure}[!ht]
\begin{center}
\includegraphics[width=3.0in]{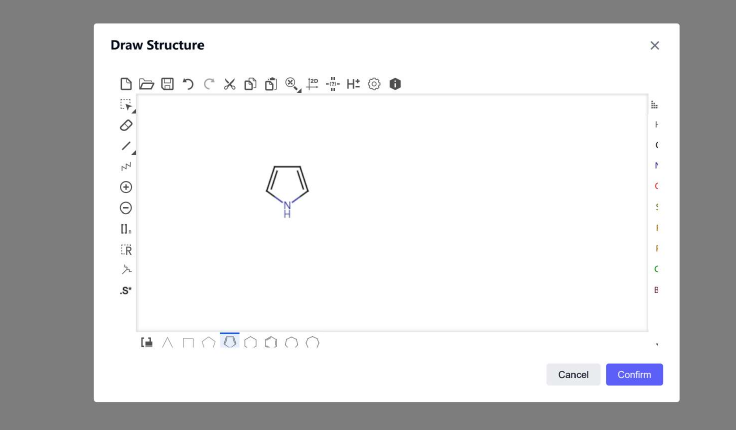}
\end{center}
\caption{Compound Structure Editor} 
\label{fig:Editor}
\end{figure}

\begin{figure}[!h]
\begin{center}
\includegraphics[width=3.0in]{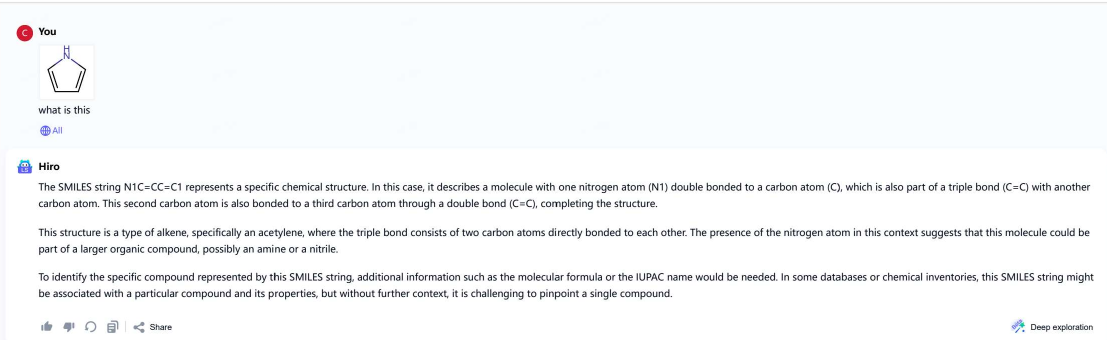}
\end{center}
\caption{Compound Structure Question Answering} 
\label{fig:Structure}
\end{figure}

As illustrated in Figures~\ref{fig:Editor} and \ref{fig:Structure}, when a user inputs or draws a chemical structure, the system leverages a dedicated API to recognize the compound. The system then retrieves and synthesizes relevant compound-related data to provide a highly accurate and contextually appropriate response to the user's query. This functionality is particularly useful for chemical and pharmaceutical research, where understanding the properties and interactions of specific compounds is essential.

\subsection{Deep Research Mode}

\begin{figure*}[!ht]
\begin{center}
\includegraphics[width=6.6in]{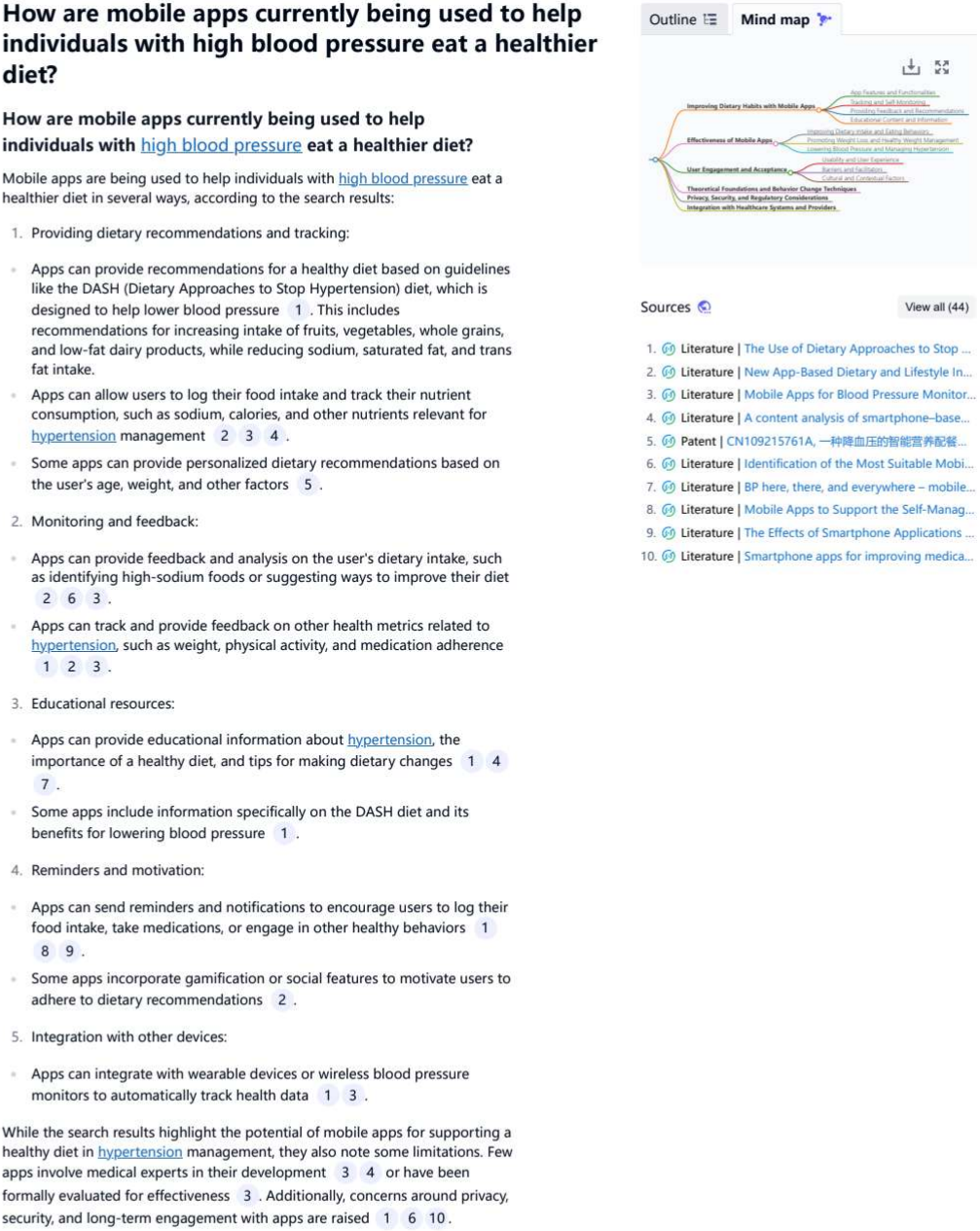}
\end{center}
\caption{Deep Research Question Answering} 
\label{fig:Deep}
\end{figure*}

For more in-depth analysis, users can activate the \textbf{Deep Research Mode}, as illustrated in Figures~\ref{fig:Deep} and \ref{fig:Graph} (Appendix~\ref{sec:appendix}). In cases where the standard response from SynapseChat lacks the necessary detail for complex research inquiries, this mode allows the system to retrieve a more comprehensive dataset, providing a thorough and integrated answer. Additionally, the system automatically generates a mind map based on the retrieved data, visualizing the relationships between key entities and concepts. This feature enhances the user's ability to understand and explore the data, facilitating deeper insights and discoveries in fields such as drug development, chemistry, and clinical research.

\section{Examples and Analysis}

Table~\ref{tab:example} (Appendix~\ref{sec:appendix}) compares the performance of SynapseChat and GPT-4-turbo in the biomedical domain, specifically focusing on the clinical results of antibody-drug conjugates (ADCs) targeting gastric cancer. When provided with identical retrieved information, SynapseChat consistently demonstrates superior performance. With its enhanced domain knowledge, PharmaGPT is more effective at selecting data relevant to user inquiries. This improvement can be attributed to its specialized knowledge, enabling it to identify and utilize the most pertinent data more effectively, resulting in more accurate and contextually appropriate responses. This capability underscores SynapseChat's potential to deliver high-quality answers in specialized biomedical research environments.

\section{Ethical Considerations}

The deployment of large language models in the life sciences raises several ethical concerns. A primary issue is data privacy, especially when handling sensitive patient information that is essential for training these models. Ensuring data security and anonymity is crucial, given the serious implications a breach could have on individual privacy and research integrity. Furthermore, using large language models to synthesize new chemical entities or predict drug interactions requires rigorous validation to ensure reliability.

\textbf{Data Privacy and Security}: Implement advanced encryption, access controls, and differential privacy to protect sensitive user data.

\textbf{Reliability and Validation of Predictions}: Ensure rigorous validation of PharmaGPT across diverse datasets and maintain transparency about model limitations to prevent misuse.

\textbf{Equitable Access}: Partner with realated organizations and consider tiered pricing or open-source licensing to facilitate broader access, especially in low-resource settings.

These strategic measures can help mitigate risks and enhance the responsible deployment of PharmaGPT in the life sciences, ensuring safety, equity, and sustainability.

\section{Conclusion}
\label{sec:conclusion}

In this work, we presented \textbf{PharmaGPT}, a foundational large language model (LLM) specifically trained on biomedical domain knowledge. PharmaGPT is integrated into a multi-channel information retrieval system that seamlessly combines data from various sources, including proprietary databases on pharmaceuticals, research papers, patents, compounds, and clinical trials. By leveraging API calls and an advanced human-computer interaction (HCI) framework, the system provides users with a highly efficient tool for addressing complex research queries.

Our system delivers rapid and contextually accurate responses in specific biomedical domains, supported by reliable references. For users seeking more detailed or exploratory research, the system can generate hierarchical responses enriched with comprehensive content. This includes automated outlines and mind maps that visualize the relationships between key entities and concepts. Such functionality significantly enhances the system's utility for researchers, clinicians, and industry professionals who depend on timely and precise information to make informed decisions.

The comparative analysis (Table~\ref{tab:example}) demonstrates that \textbf{SynapseChat}, built upon PharmaGPT, significantly outperforms general-purpose models like GPT-4-turbo in domain-specific tasks. This is particularly evident in its ability to discern and apply the most relevant data from biomedical datasets. This advantage highlights the critical importance of domain specialization in large language models, especially in fields that require high levels of precision and expertise, such as biomedical research.

In summary, \textbf{PharmaGPT} represents a significant advancement in the application of large language models to the biomedical field, providing a robust and flexible solution for domain-specific question answering and research support. Future work will focus on expanding the model's knowledge base and enhancing its capabilities in other specialized areas of biomedicine.

\ignore{
\section*{Acknowledgments}

This document has been adapted by Emily Allaway from the instructions for earlier ACL and NAACL proceedings, including those for NAACL 2024 by Steven Bethard, Ryan Cotterell and Rui Yan,
ACL 2019 by Douwe Kiela and Ivan Vuli\'{c},
NAACL 2019 by Stephanie Lukin and Alla Roskovskaya,
ACL 2018 by Shay Cohen, Kevin Gimpel, and Wei Lu,
NAACL 2018 by Margaret Mitchell and Stephanie Lukin,
Bib\TeX{} suggestions for (NA)ACL 2017/2018 from Jason Eisner,
ACL 2017 by Dan Gildea and Min-Yen Kan,
NAACL 2017 by Margaret Mitchell,
ACL 2012 by Maggie Li and Michael White,
ACL 2010 by Jing-Shin Chang and Philipp Koehn,
ACL 2008 by Johanna D. Moore, Simone Teufel, James Allan, and Sadaoki Furui,
ACL 2005 by Hwee Tou Ng and Kemal Oflazer,
ACL 2002 by Eugene Charniak and Dekang Lin,
and earlier ACL and EACL formats written by several people, including
John Chen, Henry S. Thompson and Donald Walker.
Additional elements were taken from the formatting instructions of the \emph{International Joint Conference on Artificial Intelligence} and the \emph{Conference on Computer Vision and Pattern Recognition}.
}

\newpage
\bibliography{custom}

\clearpage

\appendix

\section{Appendix}
\label{sec:appendix}

\begin{table*}[!ht]
\centering
\resizebox{\linewidth}{!}{
\begin{tabular}{c|c}
        \hline
        Question & The clinical results of ADC drugs targeting gastric cancer.  \\ \hline
        Reference[1] & [1] \{"clinical register number": "NCT04152499", "clinical title": "Abstract CT038:  \\ & Preliminary efficacy and safety results of anti-TROP2 ADC SKB264 (MK-2870) in patients (pts) with previously \\ &treated advanced gastric (G) or gastroesophageal junction (GEJ) cancer from a Phase 2 study", "phase": \\ &"Phase 2 Clinical", "conclusion": "The preliminary data suggests that pts with heavily pre-treated advanced\\ & G/GEJ cancer could achieve durable response and potentially prolonged OS from SKB264 monotherapy, \\ &with a manageable safety profile.", "drug name": "Sacituzumab tirumotecan", "indications": \\ &"Gastrooesophageal junction cancer", "target": "Tumor-associated calcium signal transducer 2", "mechanism of \\ &action": "Tumor-associated calcium signal transducer 2 modulator", "drug type": "Antibody drug conjugate (ADC)"\}  \\ 
        Reference[2] & [2] \{"clinical register number": "NCT04280341", "clinical title": "Disitamab vedotin (RC48) plus \\ & toripalimab for HER2-expressing advanced gastric or gastroesophageal junction and other solid tumours: a \\ & multicentre, open label, dose escalation and expansion phase 1 trial.", "phase": "Phase 1 Clinical", \\ & "conclusion": "Our findings suggested that RC48 plus toripalimab had a manageable safety profile and showed \\ & encouraging efficacy in pretreated patients with HER2-positive and low HER2-expressing G/GEJ cancer. The \\ & findings of our phase 1 clinical trial support further investigation of HER2-targeted ADC plus \\ & immunotherapy in HER2-expressing G/GEJ cancer and pancancer treatment in the future.", "drug name": \\ & "Disitamab Vedotin", "indications": "HER2-expressing Gastroesophageal Junction Adenocarcinoma", "target": \\ & "Tubulin, Receptor tyrosine-protein kinase erbB-2", "mechanism of action": "Microtubule inhibitor, receptor \\ & protein tyrosine kinase erbB-2 antagonist", "drug type": "Antibody drug conjugate (ADC)"\}\\
Reference[3] & [3] \{"clinical title": "Efficacy and safety of trastuzumab deruxtecan and nivolumab as third- or later-line treatment for \\ & HER2-positive advanced gastric cancer: A single-institution retrospective study.", "phase": "Not Applicable", "conclusion": \\ & "Conclusions:T-DXd and nivolumab had distinctive efficacy and toxicity profiles as third- or later-line treatment for HER2-\\ & positive AGC. Considering the distinct features of each regimen might help clinicians personalize the optimal treatment \\ & approaches for patients with HER2-positive AGC.", "drug name": "Fam-trastuzumab deruxtecan-NXKI", "indications": "HER2-\\ & positive gastric cancer", "target": "Receptor tyrosine-protein kinase erbB-2, DNA topoisomerase I", "mechanism of \\ & action": "DNA topoisomerase I inhibitor, receptor protein tyrosine kinase erbB-2 antagonist", "drug type": "Antibody drug  conjugate (ADC)"\} \\
Reference[4] & [4] \{"clinical register number": "NCT04379596", "clinical title": "Phase 1b/2, open-label dose-escalation and -expansion \\ & study evaluating trastuzumab deruxtecan (T-DXd) monotherapy and combinations in patients (pts) with HER2+ and HER2-low \\ & gastric cancer (GC): DESTINY-Gastric03 (DG-03).", "phase": "Phase 1/2 Clinical", "drug name": "Fam-trastuzumab deruxtecan-\\ & NXKI", "indications": "HER2-Low Gastric Cancer", "target": "Receptor tyrosine-protein kinase erbB-2, DNA topoisomerase I", \\ & "mechanism of action": "DNA topoisomerase I inhibitor, receptor protein tyrosine kinase erbB-2 antagonist", \\ & "drug type": "Antibody drug conjugate (ADC)"\}\\
Reference[5] & [5] \{"clinical register number": "NCT05034887", "clinical \\ & title": "Phase 2 study of trastuzumab deruxtecan as neoadjuvant treatment for HER2-positive gastric and gastroesophageal \\ & junction adenocarcinoma (EPOC2003).", "phase": "Phase 2 Clinical", "conclusion": "T-DXd monotherapy showed modest single \\ & agent activity for locally advanced HER2-positive gastric or GEJ adenocarcinoma in this phase 2 study. An additional \\ & cohort combining perioperative T-DXd with capecitabine and durvalumab is planned, to assess whether treatment efficacy and \\ & outcomes can be enhanced.", "drug name": "Fam-trastuzumab deruxtecan-NXKI", "indications": "stomach adenocarcinoma, HER2 \\ & positive Gastroesophageal Junction Adenocarcinoma", "target": "Receptor tyrosine-protein kinase erbB-2, DNA \\ & topoisomerase I", "mechanism of action": "DNA topoisomerase I inhibitor, receptor protein tyrosine kinase erbB-2 antagonist",\\ &  "drug type": "Monoclonal antibody, Antibody drug conjugate (ADC)"\}
         \\ \hline
        gpt4-turbo & In the study of the clinical results of ADC (antibody-drug conjugate) drugs targeting gastric  \\ & cancer, several drugs have shown varying degrees of efficacy and safety. The relevant clinical results of these drugs are detailed below: \\ & 

1. **Sacituzumab tirumotecan** [1]: This drug targets the TROP2 receptor and is used for the treatment of heavily pre-treated  \\ & advanced gastric cancer (G) or gastroesophageal junction cancer (GEJ). Preliminary data from a Phase 2 clinical study  \\ & indicate that monotherapy with SKB264 (the clinical name for Sacituzumab tirumotecan) can lead to durable responses and  \\ & potentially prolonged overall survival (OS) with a manageable safety profile. \\ & 

2. **Disitamab Vedotin** [2]: In a Phase 1 clinical trial, Disitamab Vedotin combined with toripalimab was used for the  \\ & treatment of HER2-expressing advanced gastric or gastroesophageal junction cancer, as well as other solid tumors.  \\ & This combination therapy showed a manageable safety profile and encouraging efficacy in previously treated HER2-positive and low HER2-expressing G/GEJ cancer patients. \\ & 

3. **Fam-trastuzumab deruxtecan-NXKI** [3][4][5]: Multiple studies on HER2-positive advanced gastric cancer have shown that \\ &  both monotherapy and combination therapy with Fam-trastuzumab deruxtecan-NXKI exhibit good efficacy and acceptable  \\ & safety profiles. Particularly in HER2-positive and HER2-low expressing gastric cancer patients, it has shown promising \\ &  clinical effects in various treatment stages, including third-line or later treatments and perioperative therapy. \\ & 

These study results not only demonstrate the potential of ADC drugs in the treatment of gastric cancer but also emphasize the  \\ & importance of continued research and development of ADC drugs targeting different receptors and combining different  \\ & therapeutic mechanisms. ADC drugs, by linking antibodies with efficacious drugs (such as chemotherapy agents), can more  \\ & precisely deliver the drugs to cancer cells, reducing the impact on normal cells, thereby enhancing efficacy and minimizing side effects.  \\ \hline
        pharma-gpt & The clinical results of ADC drugs targeting gastric cancer mainly come from the following studies:\\ &

1. **Preliminary efficacy and safety results of SKB264 (MK-2870)**:
   - **Clinical Study**: CT038\\ &
   - **Clinical Results**: Preliminary data show that patients with heavily pre-treated advanced G/GEJ cancer can achieve \\ &durable responses and potentially prolonged OS with SKB264 monotherapy, with a manageable safety profile [1].\\ &

2. **Study of Disitamab vedotin (RC48) combined with toripalimab**:
   - **Clinical Study**: NCT04280341\\ &
   - **Clinical Results**: The results indicate that RC48 combined with toripalimab shows encouraging efficacy and a \\ &manageable safety profile in HER2-expressing advanced G/GEJ cancer patients. This finding supports further research into\\ & the use of HER2-targeted ADC combined with immunotherapy in G/GEJ cancer [2].\\ &

3. **Studies of Trastuzumab deruxtecan (T-DXd)**:
   - **Clinical Study**: NCT04379596 (DESTINY-Gastric03)\\ &
   - **Clinical Results**: T-DXd shows certain efficacy in gastric cancer patients with low HER2 expression. Ongoing \\ &research aims to evaluate the combination effects of T-DXd with other treatment methods [4].\\ &
   - **Clinical Study**: NCT05034887 (EPOC2003)
   - **Clinical Results**: T-DXd monotherapy shows moderate \\ &single-agent activity in locally advanced HER2-positive gastric or GEJ adenocarcinoma. Plans are underway to combine \\ &T-DXd with capecitabine and durvalumab to evaluate whether this can enhance treatment efficacy and outcomes [5].\\ &

In summary, ADC drugs have shown certain efficacy in the treatment of gastric cancer, particularly in patients with HER2 \\ &expression or HER2-positive gastric cancer. These drugs include SKB264, Disitamab vedotin, and Trastuzumab deruxtecan. \\ &However, the specific effects and safety of each drug still require further research and validation.  \\ \hline
\end{tabular}
}
\caption{Case study.}
\label{tab:example}
\end{table*}

\begin{figure}[!ht]
\begin{center}
\includegraphics[width=3.0in]{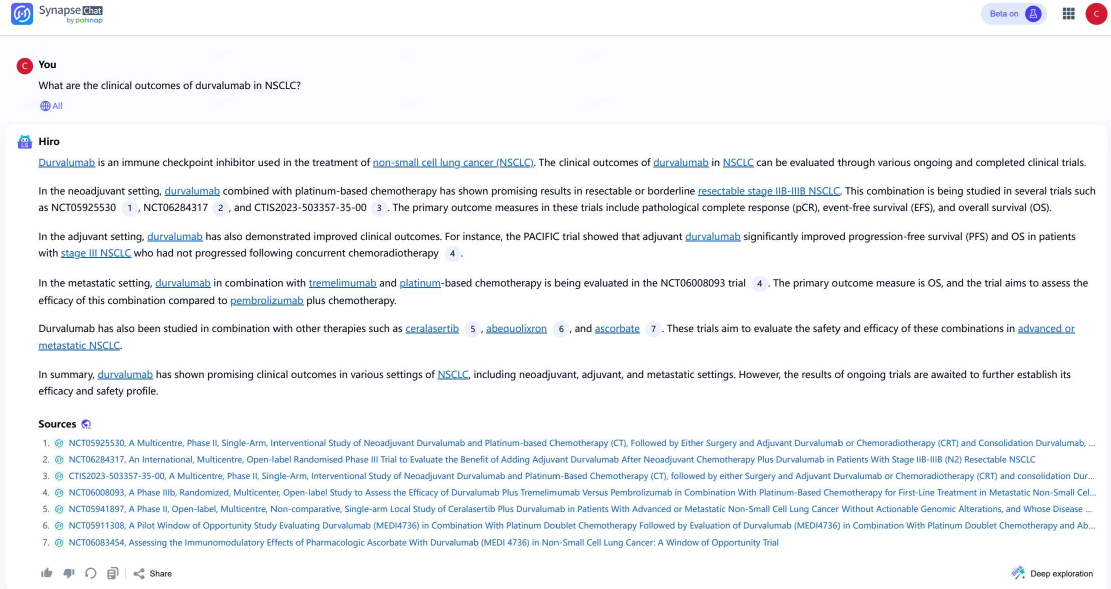}
\end{center}
\caption{Clinical Trial-related Question Answering} 
\label{fig:clinical}
\end{figure}

\begin{figure}[!ht]
\begin{center}
\includegraphics[width=3.0in]{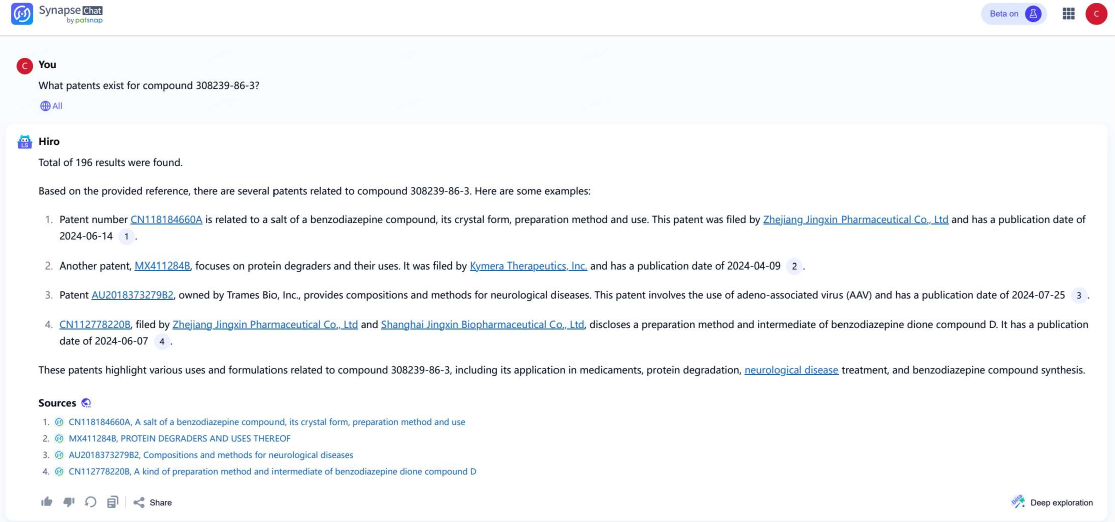}
\end{center}
\caption{Compound-related Question Answering} 
\label{fig:compound}
\end{figure}

\begin{figure}[!ht]
\begin{center}
\includegraphics[width=3.0in]{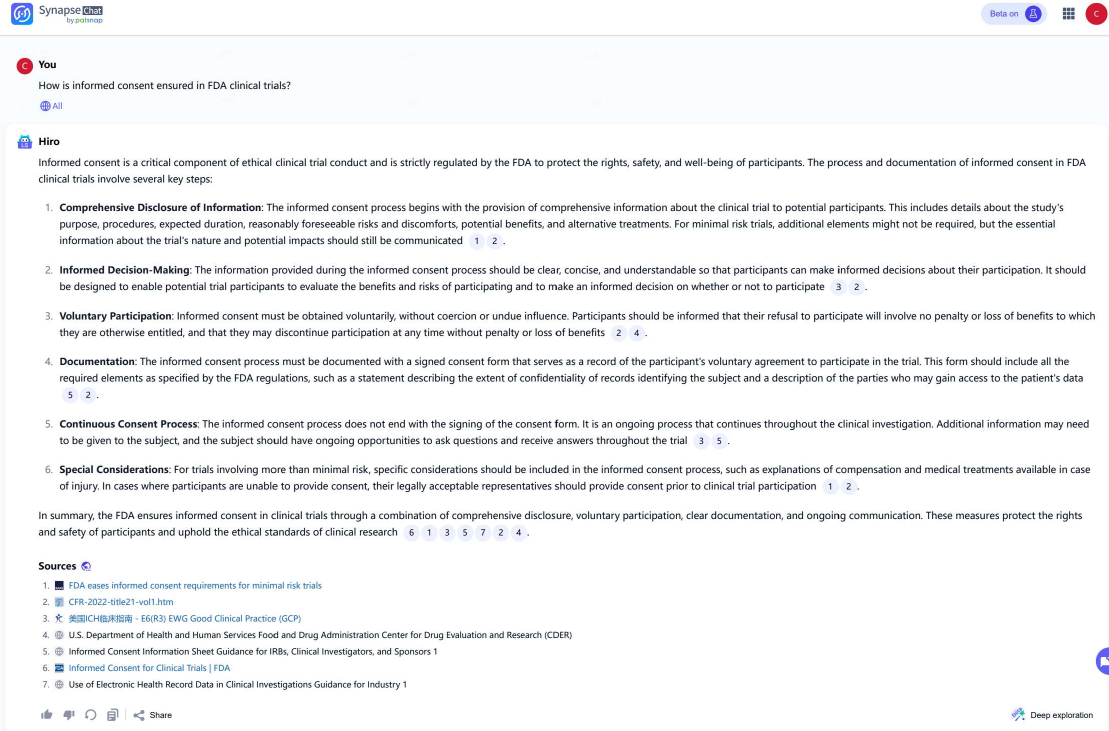}
\end{center}
\caption{Regulatory and Policy-related Question Answering} 
\label{fig:FDA}
\end{figure}

\begin{figure}[!t]
\begin{center}
\includegraphics[width=3.0in]{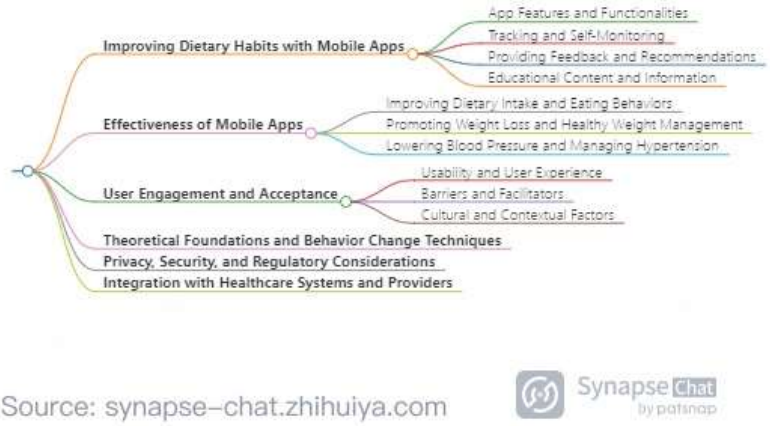}
\end{center}
\caption{Automatically Generated Mind Map} 
\label{fig:Graph}
\end{figure}

\section{Case Study: Analysis of GPT-4-turbo and PharmaGPT Responses}

In this case study (Table~\ref{tab:example}), we analyze the difference in the responses generated by GPT-4-turbo and PharmaGPT to the question: "The clinical results of ADC drugs targeting gastric cancer." Specifically, we focus on why GPT-4-turbo included Reference [3], while PharmaGPT did not, and argue that PharmaGPT provided a more accurate and professional response in this context.

\subsection{Overview of the Question and Responses}

The question asks for the clinical results of Antibody-Drug Conjugates (ADC) targeting gastric cancer. Both GPT-4-turbo and PharmaGPT provided responses summarizing the clinical results of ADC drugs such as Sacituzumab Tirumotecan, Disitamab Vedotin, and Fam-trastuzumab Deruxtecan-NXKI. While both models used References [1], [2], [4], and [5], a key difference was observed in their treatment of Reference [3].

Reference [3] describes a retrospective study involving the combination of trastuzumab deruxtecan (an ADC drug) and nivolumab (an immune checkpoint inhibitor) as a third-line or later treatment for HER2-positive advanced gastric cancer. GPT-4-turbo included Reference [3] in its response, while PharmaGPT did not. We argue that PharmaGPT's decision to exclude Reference [3] reflects a more appropriate and professional handling of the question's intent.

\subsection{Analysis of GPT-4-turbo’s Approach}

GPT-4-turbo chose to include Reference [3] in its response, likely due to its broader interpretation of the question. Since trastuzumab deruxtecan is an ADC drug, GPT-4-turbo considered the study relevant, even though it involved a combination therapy with nivolumab, an immune checkpoint inhibitor. This approach suggests that GPT-4-turbo was more inclusive, focusing on any study where ADC drugs were part of the treatment, regardless of the involvement of other therapeutic agents.

While this inclusive approach offers a broader perspective, it introduces a significant issue: the focus on ADC drugs becomes diluted by the presence of immune checkpoint inhibitors, such as nivolumab. The study in Reference [3] does not strictly isolate the effects of the ADC drug, as the outcomes could be influenced by the addition of nivolumab. Therefore, including this reference in the context of a question specifically about ADC drugs may lead to misleading conclusions regarding the efficacy and safety of the ADC alone.

\subsection{Analysis of PharmaGPT’s Approach}

In contrast, PharmaGPT's decision to exclude Reference [3] reflects a more focused and professional interpretation of the question. PharmaGPT appears to have recognized that Reference [3] deals with a combination therapy, where the effects of trastuzumab deruxtecan are intertwined with those of nivolumab, an immune checkpoint inhibitor. Given that the question explicitly asks for the clinical results of ADC drugs, PharmaGPT likely excluded this reference to provide a more accurate and relevant response focused on studies where ADC drugs were the primary or sole intervention.

By excluding Reference [3], PharmaGPT avoids the potential confusion that could arise from including a study where the therapeutic outcomes cannot be solely attributed to the ADC drug. This decision demonstrates a more nuanced understanding of the clinical trial data and a stricter adherence to the question's request for ADC-specific results. Additionally, PharmaGPT's structured response, which includes detailed references to specific clinical trials (e.g., NCT04152499, NCT04280341, NCT04379596), allows for a more precise and reliable presentation of the data.

\subsection{Professionalism and Accuracy in PharmaGPT’s Response}

PharmaGPT’s approach demonstrates a higher level of professionalism and precision for several reasons:

\begin{itemize}
    \item \textbf{Precision in Scope}: PharmaGPT interpreted the question narrowly and correctly, focusing solely on studies where ADC drugs were the primary treatment. This ensured that the results presented were directly relevant to the efficacy and safety of ADC drugs, without the confounding effects of additional therapies like nivolumab.
    
    \item \textbf{Structured and Detailed Response}: PharmaGPT provided a more structured and detailed response by clearly delineating the clinical trial results, including appropriate clinical trial identifiers (e.g., NCT numbers), target mechanisms, and study phases. This level of detail enhances the credibility of the response and allows researchers to trace the original studies easily.
    
    \item \textbf{Avoidance of Misleading Data}: By excluding Reference [3], which involved a combination therapy, PharmaGPT avoided presenting potentially misleading data that could overestimate or misattribute the efficacy of an ADC drug that was co-administered with an immune checkpoint inhibitor. This exclusion reflects a more careful and professional handling of clinical results.
\end{itemize}

\subsection{Discussion of Interpretative Differences}

The differing responses from GPT-4-turbo and PharmaGPT highlight two distinct approaches to interpreting the question:

\begin{itemize}
    \item \textbf{GPT-4-turbo's inclusive approach}: GPT-4-turbo adopted a more inclusive interpretation, allowing studies where ADC drugs were part of a combination therapy. While this approach provided a broader overview, it lacked the precision necessary to isolate the clinical effects of ADC drugs alone, potentially leading to less accurate conclusions about ADC efficacy.
    
    \item \textbf{PharmaGPT's focused and professional approach}: PharmaGPT took a more precise and conservative approach, focusing on studies where ADC drugs were the primary treatment. By doing so, PharmaGPT delivered a more accurate and relevant response tailored to the specific nature of the question, demonstrating a higher level of professionalism in clinical data interpretation.
\end{itemize}

\subsection{Conclusion}

In this case study, we demonstrated that PharmaGPT’s response was more accurate and professional compared to GPT-4-turbo’s. PharmaGPT’s decision to exclude Reference [3] reflects a more focused and precise interpretation of the question, ensuring that only relevant ADC-specific clinical results were included. Furthermore, PharmaGPT’s structured and detailed format, along with its careful selection of references, indicates a deeper understanding of the nuances of clinical trial data. In contrast, GPT-4-turbo’s broader, more inclusive approach diluted the focus on ADC drugs by including a combination therapy, which potentially misrepresents the efficacy and safety of ADC drugs alone. Therefore, PharmaGPT’s response should be considered more reliable and professional in this context.

\end{document}